\pdfoutput=1

\documentclass[11pt]{article}

\usepackage[final]{acl}

\usepackage{times}
\usepackage{latexsym}
\usepackage{array}

\usepackage[T1]{fontenc}

\usepackage[utf8]{inputenc}

\usepackage{microtype}

\usepackage{inconsolata}

\usepackage{graphicx}

\usepackage{amsmath}
\usepackage{booktabs}
\usepackage{float}
\restylefloat{table}
\usepackage{placeins}

\usepackage{cleveref} 

\newcommand{\datasetname}[1]{\texttt{#1}}
\newcommand{\modelname}[1]{\texttt{#1}}

\title{Dreaming Out Loud: A Self-Synthesis Approach For Training Vision-Language Models With Developmentally Plausible Data}

\author{
    Badr AlKhamissi\thanks{Equal Contribution} \quad
    Yingtian Tang\footnotemark[1] \quad 
    Abdülkadir Gökce\footnotemark[1] \\
    \textbf{Johannes Mehrer}\thanks{Equal Supervision} \quad 
    \textbf{Martin Schrimpf}\footnotemark[2] \\
    EPFL
}

\begin{document}
\maketitle
\begin{abstract}

    While today's large language models exhibit impressive abilities in generating human-like text, they require massive amounts of data during training. We here take inspiration from human cognitive development to train models in limited data conditions. Specifically we present a self-synthesis approach that iterates through four phases: Phase 1 sets up fundamental language abilities, training the model from scratch on a small corpus. Language is then associated with the visual environment in phase 2, integrating the model with a vision encoder to generate descriptive captions from labeled images. In the ``self-synthesis'' phase 3, the model generates captions for unlabeled images, that it then uses to further train its language component with a mix of synthetic, and previous real-world text. This phase is meant to expand the model's linguistic repertoire, similar to humans self-annotating new experiences. Finally, phase 4 develops advanced cognitive skills, by training the model on specific tasks such as visual question answering and reasoning. Our approach offers a proof of concept for training a multimodal model using a developmentally plausible amount of data.

\end{abstract}

\section{Introduction}

\begin{figure}[ht]
    \centering
    \includegraphics[width=1\linewidth]{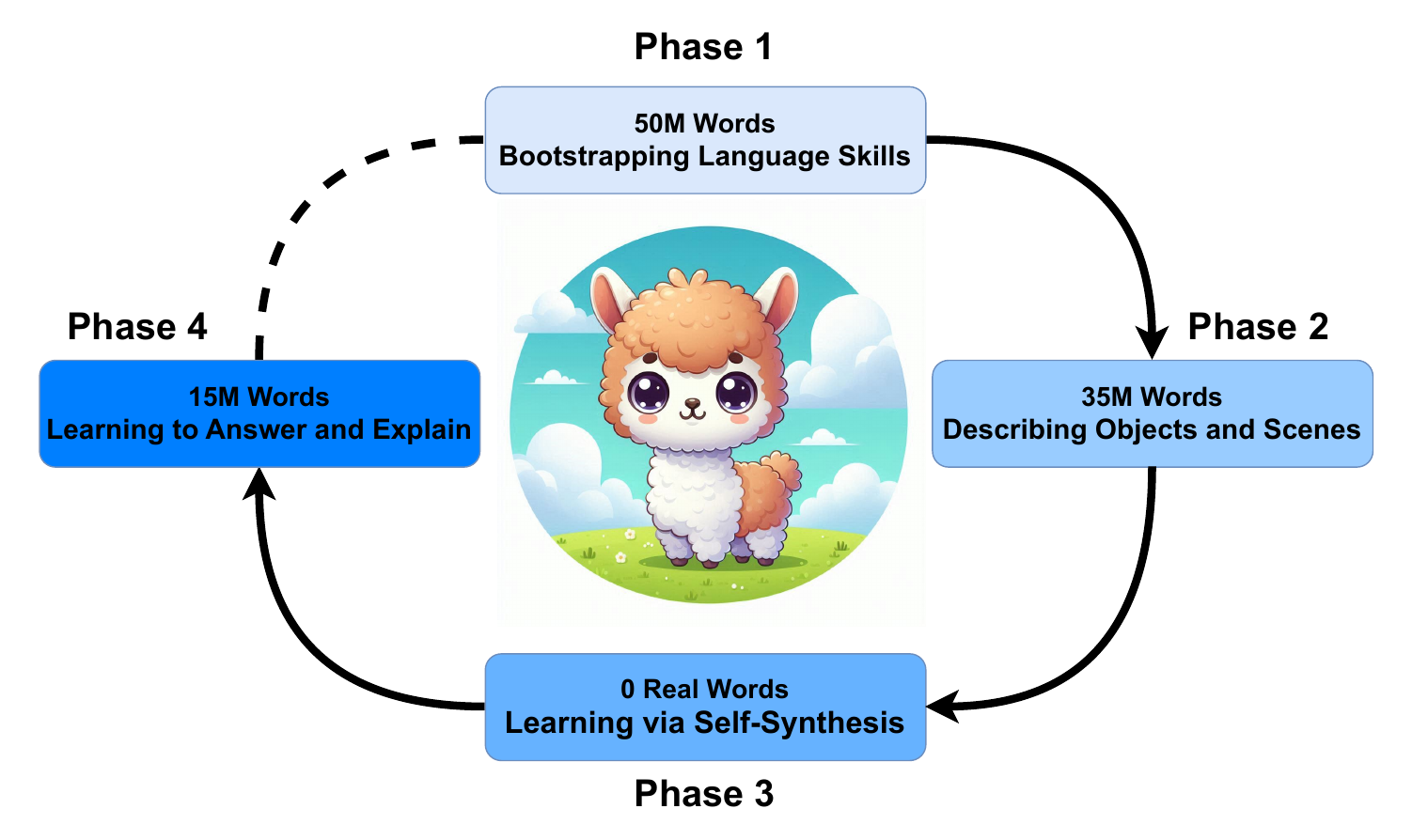}
    \caption{
        \textbf{Self-Synthesis Training Framework}. Our model \modelname{BabyLLaMA} is trained in four phases that connect fundamental language abilities to vision by learning to describe unlabeled visual experiences. We divided our approach in 4 phases, each feeding its best snapshot in terms of validation loss to the next phase. Phase 1 is concerned with fundamental language skill acquisition using 50M words. Phase 2 combines visual and text data (35 M words) to learn to describe objects and scenes. In phase 3 - making our approach one revolving around self-synthesis - we generate captions from images and use this synthesized text (i.e., 0 words from real-world corpora) to further train the model's language decoder. Phase 4 closes the loop using 15M words to develop skills for advanced visuo-linguistic tasks such as question answering and reasoning about the environment.
    }
    \label{fig:main-figure}
\end{figure}

Recent advances in machine learning have produced large language models (LLMs) that, after training on massive text corpora, are capable of generating human-like text. However, when comparing LLM training to human development, the amount of data required for successful model training far exceeds the quantities that humans learn from during their development \citep{warstadt-etal-2023-findings}. The human brain is thus often seen as a more sample-efficient learning machine than contemporary artificial neural network approaches \cite{Marcus2020TheND}.

In this work, we take inspiration from human cognitive development to build new models under limited data conditions that more closely resemble the language experience of humans. Specifically, humans learn language in combination with other senses, and use it to reflect on their experiences. We implement this idea via a \emph{self-synthesis} approach that combines vision and language such that the model learns on external (real-world) text as well as its own (synthetic) description of unlabeled visual experiences (Figure~\ref{fig:main-figure}). Self-synthesis can also be seen as analogous to the process of dreaming, which neuroscience research suggests functions as providing ``augmented samples of waking experiences,'' helping to shape neural representations and prevent overfitting to those experiences \citep{Hoel2021, prince2021}.


\section{Dataset Selection}
\label{sec:dataset}

In line with the BabyLM challenge requirements \cite{conll-2023-babylm}, we restrict our training data to 100 million words, which approximates the maximum number of words a 13-year-old would encounter in their lifetime \cite{Gilkerson2017}. In contrast, the latest \modelname{LLaMA-3-8B} model was trained on 15 trillion tokens \cite{llama3}, which is 150,000 times larger than our training budget. We created our own dataset of 100 million words, emphasizing diversity and quality. This word budget is split evenly between a text-only corpus and a multimodal image-text corpus.


\paragraph{Text-Only Data}
Our text corpus comprises 50 million words selected from the top-scoring sentences of \datasetname{FineWeb-Edu}'s October 2024 CommonCrawl snapshot \citep{lozhkov2024fineweb-edu}, based on their educational quality. \datasetname{FineWeb-Edu} is a subset of the \datasetname{FineWeb} dataset \citep{penedo2024fineweb}, which is created using scalable, automated annotations to assess educational value. The educational scores were assigned by \modelname{LLaMA-3-70B-Instruct}, which rated 500,000 samples on a scale from 0 to 5 for their educational quality \cite{penedo2024fineweb}. Models trained on this dataset have surpassed all other publicly available web datasets on several educational benchmarks, including MMLU \citep{hendrycks2021mmlu}, ARC \citep{Clark2018arc}, and OpenBookQA \citep{mihaylov-etal-2018-openbookqa}.

\paragraph{Image-Text Data}
Our image-text corpus consists of two groups: (1) image-caption data used for visual experience training (``phase 3'' \Cref{sec:phase3}); (2) multi-task image-text data used for finetuning the model towards advanced reasoning (``phase 4'', \Cref{sec:phase4}), which include captioning, VQA, and visual reasoning. For the images with captions used for visual experience training, we select subsets from \datasetname{WIT} \cite{srinivasan2021wit}, \datasetname{obelics} \cite{laurenccon2024obelics}, and \datasetname{LAION} \cite{schuhmann2021laion}.
These datasets include diverse image descriptions such as wikipedia paragraphs, news, and also simple short captions.
We sampled 27 million, 5 million, and 3 million words respectively from the 3 datasets. For the multi-task image-text data, we used \datasetname{M3IT} \cite{li2023m}, a dataset curated for multi-lingual instruction tuning and sampled 15 million words from it.
The goal is to enhance the model's ability to follow instructions as well as gain more advanced skills such as visual-reasoning, such that it can utilize its acquired knowledge more effectively.
Taken together, the two groups of image-text data make up a total of 50 million words. The selection of these datasets was not arbitrary; it resulted from multiple iterations aimed at ensuring both diversity and quality.




\section{Benchmarks}

We evaluate our model across six benchmarks: three focused on language-only tasks and three on vision-language tasks. Except for GLUE, where we fine-tune the model on each subtask using LoRA \cite{hu2022lora}, all benchmarks are evaluated in a zero-shot setting.

\subsection{Language-Only Benchmarks}

\paragraph{BLiMP}
BLiMP is a benchmark that evaluates key grammatical phenomena in English. It is composed of 67 sub-datasets, each containing 1,000 minimal pairs designed to highlight specific contrasts in syntax, morphology, or semantics. The data is automatically generated based on grammars developed by experts \cite{Warstadt2019BLiMPTB}.

\paragraph{Elements of World Knowledge (EWoK)}
EWoK is a benchmark that evaluates the world modeling abilities of language models. It covers 11 key domains of world knowledge essential for human-like world modeling. These domains range from reasoning about spatial relations to understanding social interactions \cite{ivanova2024elements}. 

\paragraph{GLUE}
The General Language Understanding Evaluation (GLUE) benchmark is a comprehensive suite of resources designed to train, evaluate, and analyze natural language understanding models. It includes nine diverse tasks focused on sentence or sentence-pair understanding, drawn from well-established datasets. These tasks vary in dataset size, text genre, and complexity, providing a broad assessment of language understanding capabilities \cite{wang-etal-2018-glue}. In our experiments, we utilize LoRA \cite{hu2022lora}, a parameter efficient finetuning method, in order to tune our model to the GLUE tasks.

\begin{figure*}[ht!]
    \centering
    \includegraphics[width=1\linewidth]{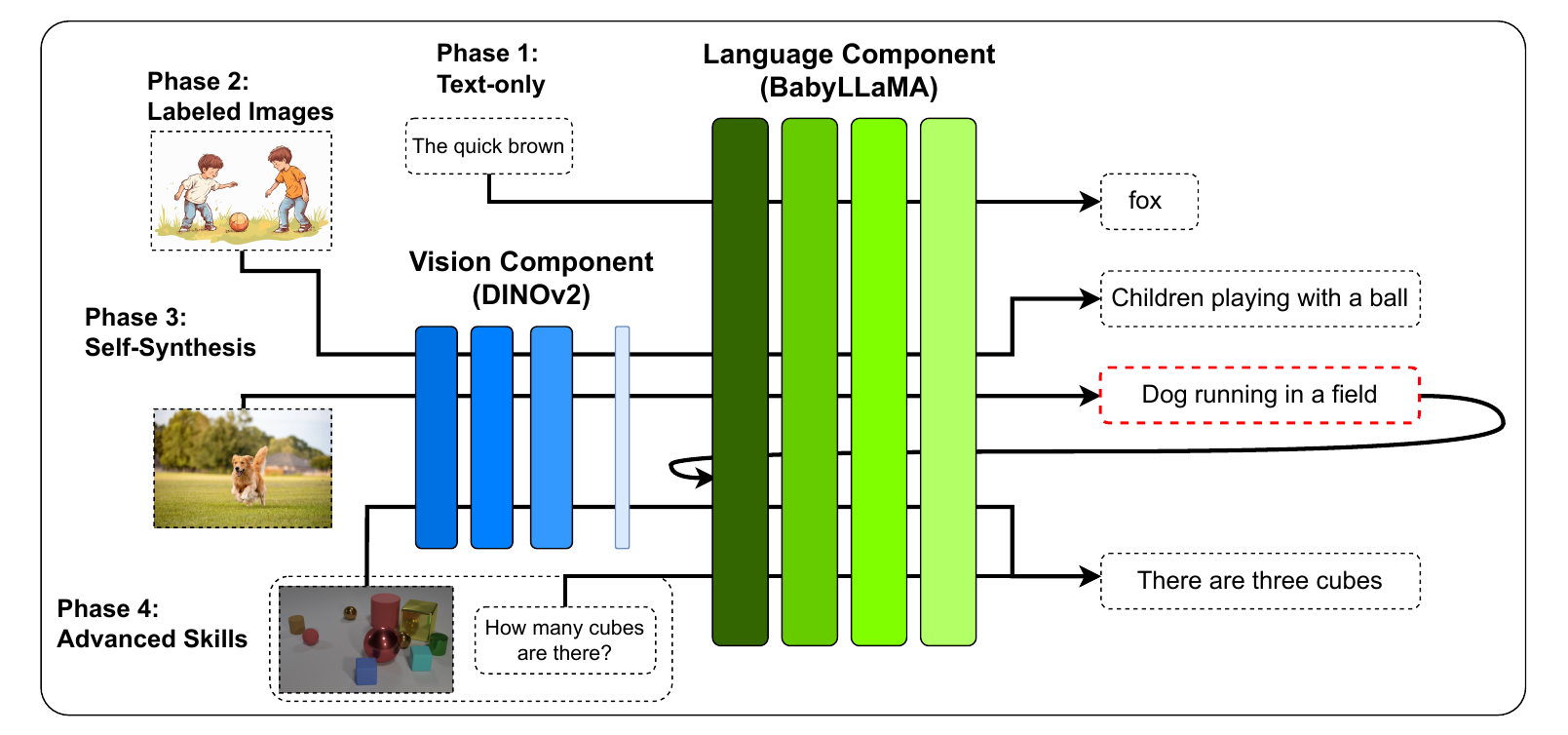}
    \caption{Overview diagram illustrating the four phases of training. Starting from training on text only (phase 1), language capabilities are connected to images (phase 2). The model then self-synthesizes text (red border) on unseen images, and uses this text to continue training the language component (phase 3), which is further refined for e.g. question answering (phase 4). Sizes of model components do not reflect number of parameters.}
    \label{fig:architecture}
\end{figure*}

\subsection{Vision-Language Benchmarks}

\paragraph{VQA}
We use the second version of the Visual Question Answering (VQA) benchmark that builds upon the original VQA \citep{vqav1} by incorporating complementary images. In this dataset, each question is linked to a pair of similar images, each yielding a distinct answer, thus increasing the challenge. For the model to answer these questions, it requires a grasp of vision, language, and commonsense knowledge \citep{vqav2}.

\paragraph{Winoground}
Winoground is a challenging task and dataset designed to assess the visio-linguistic compositional reasoning abilities of vision-language models. The objective is to correctly match two images with two captions, where both captions use the exact same words or morphemes but arranged in different orders. Expert annotators carefully curated the dataset, providing fine-grained tags to facilitate a detailed analysis of model performance \cite{thrush_and_ross2022winoground}.

\paragraph{DevBench}
This benchmark contains 7 tasks across lexical, syntactic, and semantic domains, each accompanied by human response data at the item level, allowing for detailed comparisons between model scores and human response distributions. The lexical tasks evaluate vocabulary knowledge by assessing the model's ability to correctly identify the visual referent of a given noun. Syntactic tasks test grammatical understanding, requiring the model to choose the correct scene that aligns with a provided sentence. Semantic tasks measure the model’s ability to represent conceptual similarity, either visually or linguistically, by comparing representational similarity scores \cite{Tan2024DevBenchAM}.

\section{Model Details}


We use the same model architecture provided by the BabyLM Challenge organizers, called \modelname{BabyLLaMA}, which consists of a reduced \modelname{LLaMA} architecture, and we equip it with the \modelname{DINOv2}$_{\text{Large}}$ vision encoder to be able to handle visual inputs.

\paragraph{Tokenizer} 
We train a BPE tokenizer with a vocabulary size of 16,000 on the text data from the curated dataset described in Section \ref{sec:dataset}.

\paragraph{Language Model} 
The language models employs the same architectural components as the \modelname{LLaMA} model \cite{llama3}, but with only 16 Transformer layers and a reduced hidden dimension size of 512. The intermediate size in the MLP is 1,024, resulting in a total model size of 58 million parameters. The attention mechanism uses 8 attention heads in each layer.

\paragraph{Vision Encoder}
To equip our language model with visual capabilities, we incorporate the pretrained \modelname{DINOv2}$_{\text{Large}}$ vision encoder \cite{Oquab2023DINOv2LR}. \modelname{DINOv2} was trained on a large-scale, unlabeled image dataset. It is built on the ViT architecture \cite{vit} and generates 256 vision tokens per image. The vision encoder remains frozen during all experiments to preserve its pretrained features. 

\begin{table*}[htpb!]
\centering
\begin{tabular}{cp{0.7\linewidth}}
\toprule
\textbf{Image} & \textbf{Synthetic Description} \\
\midrule
\includegraphics[width=3cm]{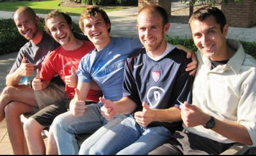} & \vspace{-1.7cm} Hot off the field at the Ravensboro Golf Club in Ravensboro, IL. I am looking forward to the win of the season. I love the game and hopefully the games are really going to be a big thing... \\
\includegraphics[width=3cm]{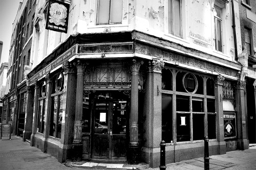} & \vspace{-1.7cm} The first section of the East End of London's West End was dedicated to the Holy Spirit. The West End of London's West End was the last part of the East End of London... \\
\includegraphics[width=3cm]{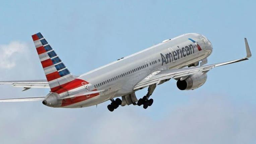} & \vspace{-1.7cm} The airport is in the midst of a multi-year, \$10 billion contract with the U.S. Navy, which is expected to be operational over the next few years. The agreement is expected to be signed by the United States, Canada, and the United States... \\
\bottomrule
\end{tabular}
\caption{Synthetic descriptions generated by the model for the images shown. This table illustrates the model's ability to associate visual cues with corresponding textual representations.}
\label{tab:synthetic_desc}
\end{table*}

\paragraph{Projection Module}
The projection module serves as the bridge between the vision encoder and the language model. It comprises a two-layer MLP with a GeLU activation function in between. This module projects the concatenated image tokens to match the dimensionality of the language model and is learnable throughout the training process.

\section{Self-Synthesis Training Phases}

Our framework trains the model in four phases. In each phase, we record the model checkpoint with the lowest validation loss and use it as a starting point for the following phase. For all phases, we use the AdamW optimizer combined with a cosine learning rate scheduler and a batch-size of 256. The learning rate begins with a linear warm-up phase and then gradually decreases to zero over the course of the training.

\subsection{Phase 1: Bootstrapping Language Skills}
\label{sec:phase1}

Similar to how children learn a fundamental linguistic repertoire with supervision from their environment, the language component of our model is first trained from scratch on a text-only corpus. Specifically, we train \modelname{BabyLLaMA} for 15 epochs on fewer than 50 million words, using the top-scoring sentences from \datasetname{FineWeb-Edu} based on their educational quality. Rather than concatenating and chunking the entire corpus into the maximum sequence length, as is common in language model pretraining, we divided each document from the \datasetname{FineWeb-Edu} snapshot into individual sentences. Each sentence was truncated to have a maximum of 256 tokens and a minimum of 10 tokens. We found that training on shorter sequences by segmenting documents in this way resulted in better performance on the \datasetname{BLiMP} benchmark \citep{Warstadt2019BLiMPTB} compared to training with fixed long sequences. The model was trained with a peak learning rate of $1e-4$ and a linear warm-up for the first $5,000$ optimization steps. (Learning rates $1e-4,5e-5,1e-5$ were tried and the one with the lowest validation error was chosen. We did not conduct other hyperparameter selections due to the limited resources. This also applies to other phases.)



\begin{table*}
\centering
\begin{tabular}{l|cccc|cccc}
\toprule
& \multicolumn{4}{c|}{\textit{Language-Only Benchmarks}} & \multicolumn{3}{c}{\textit{Vision-Language Benchmarks}} & \\
\textbf{Phase} & \textbf{BLiMP} & \textbf{BLiMP Supp.} & \textbf{EWoK} & \textbf{GLUE} & \textbf{VQA} & \textbf{Winoground} & \textbf{DevBench} \\
\midrule
\textbf{Phase 1} & 0.723 & 0.533 & 0.500 & 0.651     & -     & -     & -   \\
\textbf{Phase 2} & 0.728 & \textbf{0.561} & 0.504 & 0.650     & 0.395 & 0.507 & 0.242  \\
\textbf{Phase 3} & \textbf{0.736} & 0.556 & \textbf{0.514} & 0.647 & 0.380     & 0.507     & \textbf{0.350}    \\
\textbf{Phase 4} & 0.729 & 0.542 & 0.502 & \textbf{0.659} & \textbf{0.420} & \textbf{0.509} & 0.228 \\
\bottomrule
\end{tabular}
\caption{Performance comparison of the model across different phases of training on various benchmarks. The results show accuracy scores on language-only benchmarks (BLiMP, BLiMP Supp., EWoK, GLUE) and multimodal tasks (VQA, Winoground, DevBench). All benchmarks are evaluated in a zeroshot manner, except for GLUE, which is first finetuned using LoRA for each of its tasks separately. The best result across phases is highlighted in \textbf{bold}.}

\label{tab:results}
\end{table*}

\subsection{Phase 2: Learning to Associate Language and Vision}
\label{sec:phase2}

Inspired by children learning to associate words with the objects they encounter daily, this training phase integrates a \modelname{DINOv2}$_{\text{Large}}$ vision encoder into the model to link visual inputs with language.
The model is trained on image-text pairs, keeping the weights of the vision encoder frozen. We first divide each image into 16x16 patches. These $256$ tokens are then transformed into feature embeddings by the model. We concatenate every 4 consecutive tokens together to form one embedding to reduce the number of tokens from 256 to 64 before passing them to the projection module. Training involves an autoregressive loss applied exclusively to the text tokens, conditioned on the corresponding image embeddings. In this setup, the projected image embeddings are concatenated with the text embeddings \(\mathbf{t}_{1:s}\) before being passed through the language model. This allows the model to learn a joint representation that conditions the text generation on the visual context provided by the image. 

Formally, let \(\mathbf{i} = \{i_1, i_2, \dots, i_{64}\}\) be the set of image embeddings produced by the vision encoder for a given image, and \(\mathbf{t} = \{t_1, t_2, \dots, t_s\}\) be the sequence of text tokens associated with that image, where \(s \leq 512\). The training objective is to maximize the conditional likelihood of the next text token \(t_{s+1}\) given the projected image embeddings and the preceding text tokens, where $f$ is the projection module. This can be formulated as:

\begin{equation*}
   \max_{\theta, \phi} \sum_{n=1}^{N} \sum_{s=1}^{|\mathbf{t}_n|} \log p_{\theta, \phi} \left( ~t_{n, s+1} \mid [~ f(\mathbf{i}_n); \mathbf{t}_{n, 1:s}~] \right)    
\end{equation*}

where: \(p_{\theta, \phi}(\cdot)\) is the probability distribution generated by the combined model, \(f(\mathbf{i}_n)\) represents the image embeddings processed through the projection module, \(\mathbf{t}_n = \{t_1, t_2, \dots, t_s\}\) are the text tokens for the \(n\)-th image-text pair, \(N\) is the total number of training examples, and \(|\mathbf{t}_n|\) is the length of the \(n\)-th text sequence.


Therefore, just as children learn to describe their visual environment based on supervisory signals (e.g. parents describing the surroundings), the model learns to generate captions for images, articulating what it ``sees.'' To achieve this, we train the model to produce detailed descriptions across a diverse range of images. Consequently, we balanced the datasets to include samples with detailed descriptions (from \datasetname{WIT} and \datasetname{obelics}; 35842 samples / 6M words, 135393 samples / 21M words
) alongside those with concise captions (from \datasetname{LAION}; 323929 samples / 3M words). It is worth noting that although \datasetname{LAION} contains only 3 million words, it accounts for more than half of the images due to its short captions. In this phase, we train the model for $5$ epochs, with a learning rate that linearly warms-up to $10^{-5}$ for $250$ steps, then decreases to zero throughout training.




\subsection{Phase 3: Learning via Self-Synthesis}
\label{sec:phase3}

\paragraph{Self-Synthesis Using Images in the Wild.} 
Beyond supervised learning on images, children also imagine and narrate stories about what they have seen. We implement this idea by having the model generate text from a set of unlabeled images and synthesizing captions that are then used to further train the language component with more diverse text. Concretely, we collected 1.1 million images from \datasetname{obelics} that were not used during training. Using nucleus sampling (p=0.95) and top-k sampling (k=50) with a temperature of 0.7, we generated a total of 42 million words. For each image, a maximum token length between 32 and 64 was uniformly sampled. Table \ref{tab:synthetic_desc} shows a few examples of images and their corresponding text generated by our model. To avoid repetition in the generated text, we limit the maximal number of generated tokens to be 256. Note that some descriptions do not perfectly match the content of the images. This is insofar not an issue, as grammatically and vocabulary-rich text suffices for our purpose. 



\paragraph{Continuing Pretraining} 
Inspired by humans mixing real and imagined experiences to enhance their understanding, we train \modelname{BabyLLaMA} on a mixture of self-synthesized text and previously seen "real-world" data to deepen its language abilities.
Specifically, we transition back from image-text training to text-only training, combining all the text data we have gathered thus far. This results in a total of 85 million real words and 42 million synthetic words. Our model is trained for just 2 epochs, with a learning rate that linearly warms up to 1e-5 over 500 optimization steps then decreases towards zero. To assess the contribution of the self-synthesized text, we train another model version using only the 85 million real words and report the results on the text benchmarks in \Cref{sec:ablation}.

\subsection{Phase 4: Learning to Answer and Explain}
\label{sec:phase4}

Equipped with fundamental language skills and the ability to describe their surroundings, human cognitive development includes answering questions and reasoning about their environment. Similarly, we train \modelname{BabyLLaMA} to handle complex visual-linguistic tasks:
We finetune the language model along with the projection layer on \datasetname{M3IT}. We set the learning rate to $10^{-5}$ with $250$ warm-up updates. The model is trained for $2$ epochs.

The division in 4 training phases is inspired by language acquisition in human infants. However, we do not suggest that the exact same phases accurately describe human linguistic development. For example, humans are unlikely to establish fundamental language skills (phase 1) without concurrent visual input that our model only encounters in phase 2.

\begin{figure}
    \centering
    \includegraphics[width=1\linewidth]{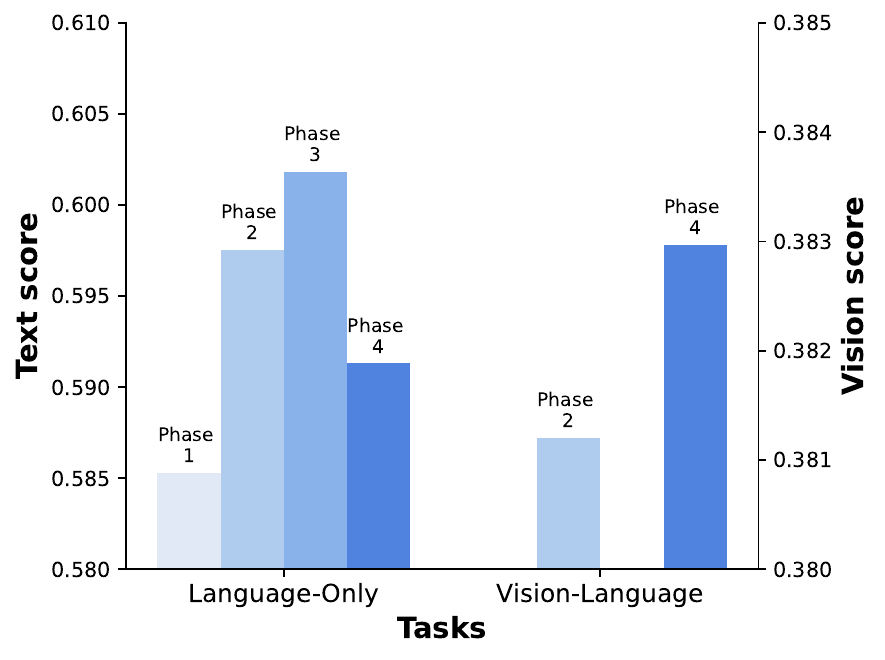}
    \caption{Average performance on all language-only (left) and vision-language-benchmarks (right) across training phases. Each phase yields a small boost for its respective training objective.}
    \label{fig:blimp}
\end{figure}

\section{Results}

\Cref{tab:results} presents the performance across various benchmarks, including both language-only and vision-language datasets. For language-only benchmarks, the phase 3 model significantly outperforms earlier models on BLiMP and EWoK, while the phase 4 model achieves the best results on GLUE. Notably, the phase 2 model delivers the highest performance on BLiMP Supplement, which is a smaller dataset compared to BLiMP.
In vision-language benchmarks, the phase 4 model surpasses the phase 3 model on VQA and Winoground but underperforms on DevBench. Overall, models after phase 3 achieve the highest scores across most benchmarks.
To emphasize performance differences across training phases, \Cref{fig:blimp} illustrates the average scores on various benchmarks. For language-only tasks, the phase 3 model shows a substantial improvement over models from phases 1 and 2. However, the phase 4 model lags slightly, likely due to fine-tuning on question-answer datasets, which shifts its focus away from general text modeling.
\Cref{tab:synthetic_desc} provides examples of synthetic descriptions generated by the phase 2 model conditioned on different images. The model accurately captures key elements in the images and produces varied syntactic and content-rich descriptions. However, there are occasional issues with logical consistency, such as the repetition of "United States" in the third example.

\subsection{Ablation Study}
\label{sec:ablation}

\begin{table}[]
\centering
\begin{tabular}{@{}lll@{}}
\toprule

\textbf{Benchmark}   & \textbf{+ Synth}        & \textbf{- Synth}        \\ \midrule
BLiMP       & \textbf{0.736} & \textbf{0.736} \\
BLiMP Supp. & \textbf{0.556} & 0.550          \\
EWoK        & \textbf{0.514} & 0.510          \\ \bottomrule
\end{tabular}
\caption{Results of the ablation study on language-only benchmarks, comparing the performance of the model trained solely on real-world text (-Synth) against the model trained on a combination of real and synthetic data (+Synth). All benchmarks were evaluated in a zero-shot manner, illustrating the contribution of synthetic data to overall model performance.}
\label{tab:ablation}
\end{table}

To measure the contribution of the synthetic data, we train a separate phase 3 model using only real-world text, excluding any generated text, and compare its performance with the model trained on a mixture of both real and synthetic data. \Cref{tab:ablation} presents the results on the language-only benchmarks, all evaluated in a zero-shot manner. The findings demonstrate that incorporating synthetic data either enhances or maintains performance across benchmarks, highlighting the potential of scaling self-synthesis with larger datasets.

\section{Conclusion}

This work proposes a novel self-synthesis approach to training vision-language models in a data-efficient manner inspired by human cognitive development. By structuring the learning process into four distinct phases---beginning with foundational language abilities, integrating vision and language, generating synthetic data through unlabeled image captioning, and advancing cognitive tasks---the resulting model is able to solve both vision-language and language only benchmarks using a limited amount of data in a unified manner.

While we observed improved performance from each phase of training, these improvements were comparatively small.
Curriculum learning methods or architectural modifications might further improve the model’s learning efficiency within the proposed framework. For instance, the phases could be ran repeatedly, such that the model iteratively trains on a mix of real-world text and continuously improving self-synthesized text. A layer-fusion approach could better utilize intermediate layer representations, which has been shown to enhance training in data-limited settings \cite{elnokrashy-etal-2024-depth}. These efforts could close the performance gap while maintaining the developmental plausibility of the training setup. In summary, results presented here suggest that self-synthesis can make effective use of information across modalities, and might help to train performant models with developmentally plausible data regimes.



\bibliography{custom}

\begin{thebibliography}{27}
\providecommand{\natexlab}[1]{#1}

\bibitem[{Clark et~al.(2018)Clark, Cowhey, Etzioni, Khot, Sabharwal, Schoenick, and Tafjord}]{Clark2018arc}
Peter Clark, Isaac Cowhey, Oren Etzioni, Tushar Khot, Ashish Sabharwal, Carissa Schoenick, and Oyvind Tafjord. 2018.
\newblock \href {https://api.semanticscholar.org/CorpusID:3922816} {Think you have solved question answering? try arc, the ai2 reasoning challenge}.
\newblock \emph{ArXiv}, abs/1803.05457.

\bibitem[{Dosovitskiy et~al.(2020)Dosovitskiy, Beyer, Kolesnikov, Weissenborn, Zhai, Unterthiner, Dehghani, Minderer, Heigold, Gelly, Uszkoreit, and Houlsby}]{vit}
Alexey Dosovitskiy, Lucas Beyer, Alexander Kolesnikov, Dirk Weissenborn, Xiaohua Zhai, Thomas Unterthiner, Mostafa Dehghani, Matthias Minderer, Georg Heigold, Sylvain Gelly, Jakob Uszkoreit, and Neil Houlsby. 2020.
\newblock \href {https://api.semanticscholar.org/CorpusID:225039882} {An image is worth 16x16 words: Transformers for image recognition at scale}.
\newblock \emph{ArXiv}, abs/2010.11929.

\bibitem[{Dubey et~al.(2024)Dubey, Jauhri, Pandey, Kadian, Al-Dahle, Letman, and et~al.}]{llama3}
Abhimanyu Dubey, Abhinav Jauhri, Abhinav Pandey, Abhishek Kadian, Ahmad Al-Dahle, Aiesha Letman, and et~al. 2024.
\newblock \href {https://api.semanticscholar.org/CorpusID:271571434} {The llama 3 herd of models}.
\newblock \emph{ArXiv}, abs/2407.21783.

\bibitem[{ElNokrashy et~al.(2024)ElNokrashy, AlKhamissi, and Diab}]{elnokrashy-etal-2024-depth}
Muhammad ElNokrashy, Badr AlKhamissi, and Mona Diab. 2024.
\newblock \href {https://aclanthology.org/2024.lrec-main.417} {Depth-wise attention ({DWA}tt): A layer fusion method for data-efficient classification}.
\newblock In \emph{Proceedings of the 2024 Joint International Conference on Computational Linguistics, Language Resources and Evaluation (LREC-COLING 2024)}, pages 4665--4674, Torino, Italia. ELRA and ICCL.

\bibitem[{Gilkerson et~al.(2017)Gilkerson, Richards, Warren, Montgomery, Greenwood, Kimbrough~Oller, Hansen, and Paul}]{Gilkerson2017}
Jill Gilkerson, Jeffrey~A. Richards, Steven~F. Warren, Judith~K. Montgomery, Charles~R. Greenwood, D.~Kimbrough~Oller, John H.~L. Hansen, and Terrance~D. Paul. 2017.
\newblock \href {https://doi.org/10.1044/2016_ajslp-15-0169} {Mapping the early language environment using all-day recordings and automated analysis}.
\newblock \emph{American Journal of Speech-Language Pathology}, 26(2):248–265.

\bibitem[{Goyal et~al.(2016)Goyal, Khot, Summers-Stay, Batra, and Parikh}]{vqav2}
Yash Goyal, Tejas Khot, Douglas Summers-Stay, Dhruv Batra, and Devi Parikh. 2016.
\newblock \href {https://api.semanticscholar.org/CorpusID:8081284} {Making the v in vqa matter: Elevating the role of image understanding in visual question answering}.
\newblock \emph{International Journal of Computer Vision}, 127:398 -- 414.

\bibitem[{Hendrycks et~al.(2021)Hendrycks, Burns, Basart, Zou, Mazeika, Song, and Steinhardt}]{hendrycks2021mmlu}
Dan Hendrycks, Collin Burns, Steven Basart, Andy Zou, Mantas Mazeika, Dawn Song, and Jacob Steinhardt. 2021.
\newblock Measuring massive multitask language understanding.
\newblock \emph{Proceedings of the International Conference on Learning Representations (ICLR)}.

\bibitem[{Hoel(2021)}]{Hoel2021}
Erik Hoel. 2021.
\newblock \href {https://doi.org/10.1016/j.patter.2021.100244} {The overfitted brain: Dreams evolved to assist generalization}.
\newblock \emph{Patterns}, 2(5):100244.

\bibitem[{Hu et~al.(2022)Hu, Shen, Wallis, Allen-Zhu, Li, Wang, Wang, and Chen}]{hu2022lora}
Edward~J Hu, Yelong Shen, Phillip Wallis, Zeyuan Allen-Zhu, Yuanzhi Li, Shean Wang, Lu~Wang, and Weizhu Chen. 2022.
\newblock \href {https://openreview.net/forum?id=nZeVKeeFYf9} {Lo{RA}: Low-rank adaptation of large language models}.
\newblock In \emph{International Conference on Learning Representations}.

\bibitem[{Ivanova et~al.(2024)Ivanova, Sathe, Lipkin, Kumar, Radkani, Clark, Kauf, Hu, RT, Grand, Paulun, Ryskina, Akyurek, Wilcox, Rashid, Choshen, Levy, Fedorenko, Tenenbaum, and Andreas}]{ivanova2024elements}
Anna Ivanova, Aalok Sathe, Benjamin Lipkin, Unnathi Kumar, Setayesh Radkani, Thomas~H Clark, Carina Kauf, Jennifer Hu, Pramod RT, Gabriel Grand, Vivian Paulun, Maria Ryskina, Ekin Akyurek, Ethan Wilcox, Nafisa Rashid, Leshem Choshen, Roger Levy, Evelina Fedorenko, Josh Tenenbaum, and Jacob Andreas. 2024.
\newblock Elements of world knowledge (ewok): A cognition-inspired framework for evaluating basic world knowledge in language models.
\newblock \emph{arXiv}.

\bibitem[{Lauren{\c{c}}on et~al.(2024)Lauren{\c{c}}on, Saulnier, Tronchon, Bekman, Singh, Lozhkov, Wang, Karamcheti, Rush, Kiela et~al.}]{laurenccon2024obelics}
Hugo Lauren{\c{c}}on, Lucile Saulnier, L{\'e}o Tronchon, Stas Bekman, Amanpreet Singh, Anton Lozhkov, Thomas Wang, Siddharth Karamcheti, Alexander Rush, Douwe Kiela, et~al. 2024.
\newblock Obelics: An open web-scale filtered dataset of interleaved image-text documents.
\newblock \emph{Advances in Neural Information Processing Systems}, 36.

\bibitem[{Li et~al.(2023)Li, Yin, Li, Chen, Wang, Ren, Li, Yang, Xu, Sun et~al.}]{li2023m}
Lei Li, Yuwei Yin, Shicheng Li, Liang Chen, Peiyi Wang, Shuhuai Ren, Mukai Li, Yazheng Yang, Jingjing Xu, Xu~Sun, et~al. 2023.
\newblock M $^3$ it: A large-scale dataset towards multi-modal multilingual instruction tuning.
\newblock \emph{arXiv preprint arXiv:2306.04387}.

\bibitem[{Lozhkov et~al.(2024)Lozhkov, Ben~Allal, von Werra, and Wolf}]{lozhkov2024fineweb-edu}
Anton Lozhkov, Loubna Ben~Allal, Leandro von Werra, and Thomas Wolf. 2024.
\newblock \href {https://doi.org/0.57967/hf/2497} {Fineweb-edu}.

\bibitem[{Marcus(2020)}]{Marcus2020TheND}
Gary~F. Marcus. 2020.
\newblock \href {https://api.semanticscholar.org/CorpusID:211126492} {The next decade in ai: Four steps towards robust artificial intelligence}.
\newblock \emph{ArXiv}, abs/2002.06177.

\bibitem[{Mihaylov et~al.(2018)Mihaylov, Clark, Khot, and Sabharwal}]{mihaylov-etal-2018-openbookqa}
Todor Mihaylov, Peter Clark, Tushar Khot, and Ashish Sabharwal. 2018.
\newblock \href {https://doi.org/10.18653/v1/D18-1260} {Can a suit of armor conduct electricity? a new dataset for open book question answering}.
\newblock In \emph{Proceedings of the 2018 Conference on Empirical Methods in Natural Language Processing}, pages 2381--2391, Brussels, Belgium. Association for Computational Linguistics.

\bibitem[{Oquab et~al.(2023)Oquab, Darcet, Moutakanni, Vo, Szafraniec, Khalidov, Fernandez, Haziza, Massa, El-Nouby, Assran, Ballas, Galuba, Howes, Huang, Li, Misra, Rabbat, Sharma, Synnaeve, Xu, J{\'e}gou, Mairal, Labatut, Joulin, and Bojanowski}]{Oquab2023DINOv2LR}
Maxime Oquab, Timoth'ee Darcet, Th{\'e}o Moutakanni, Huy~Q. Vo, Marc Szafraniec, Vasil Khalidov, Pierre Fernandez, Daniel Haziza, Francisco Massa, Alaaeldin El-Nouby, Mahmoud Assran, Nicolas Ballas, Wojciech Galuba, Russ Howes, Po-Yao~(Bernie) Huang, Shang-Wen Li, Ishan Misra, Michael~G. Rabbat, Vasu Sharma, Gabriel Synnaeve, Huijiao Xu, Herv{\'e} J{\'e}gou, Julien Mairal, Patrick Labatut, Armand Joulin, and Piotr Bojanowski. 2023.
\newblock \href {https://api.semanticscholar.org/CorpusID:258170077} {Dinov2: Learning robust visual features without supervision}.
\newblock \emph{ArXiv}, abs/2304.07193.

\bibitem[{Penedo et~al.(2024)Penedo, Kydlíček, allal, Lozhkov, Mitchell, Raffel, Werra, and Wolf}]{penedo2024fineweb}
Guilherme Penedo, Hynek Kydlíček, Loubna~Ben allal, Anton Lozhkov, Margaret Mitchell, Colin Raffel, Leandro~Von Werra, and Thomas Wolf. 2024.
\newblock \href {https://arxiv.org/abs/2406.17557} {The fineweb datasets: Decanting the web for the finest text data at scale}.
\newblock \emph{Preprint}, arXiv:2406.17557.

\bibitem[{Prince and Richards(2021)}]{prince2021}
Luke~Y. Prince and Blake~A. Richards. 2021.
\newblock \href {https://doi.org/10.1016/j.patter.2021.100268} {The overfitted brain hypothesis}.
\newblock \emph{Patterns}, 2(5):100268.

\bibitem[{Schuhmann et~al.(2021)Schuhmann, Vencu, Beaumont, Kaczmarczyk, Mullis, Katta, Coombes, Jitsev, and Komatsuzaki}]{schuhmann2021laion}
Christoph Schuhmann, Richard Vencu, Romain Beaumont, Robert Kaczmarczyk, Clayton Mullis, Aarush Katta, Theo Coombes, Jenia Jitsev, and Aran Komatsuzaki. 2021.
\newblock Laion-400m: Open dataset of clip-filtered 400 million image-text pairs.
\newblock \emph{arXiv preprint arXiv:2111.02114}.

\bibitem[{Srinivasan et~al.(2021)Srinivasan, Raman, Chen, Bendersky, and Najork}]{srinivasan2021wit}
Krishna Srinivasan, Karthik Raman, Jiecao Chen, Michael Bendersky, and Marc Najork. 2021.
\newblock Wit: Wikipedia-based image text dataset for multimodal multilingual machine learning.
\newblock In \emph{Proceedings of the 44th international ACM SIGIR conference on research and development in information retrieval}, pages 2443--2449.

\bibitem[{Tan et~al.(2024)Tan, Yu, Long, Ma, Murray, Silverman, Yeatman, and Frank}]{Tan2024DevBenchAM}
Alvin Wei~Ming Tan, Sunny Yu, Bria Long, Wanjing~Anya Ma, Tonya Murray, Rebecca~D. Silverman, Jason~D. Yeatman, and Michael~C. Frank. 2024.
\newblock \href {https://api.semanticscholar.org/CorpusID:270521851} {Devbench: A multimodal developmental benchmark for language learning}.
\newblock \emph{ArXiv}, abs/2406.10215.

\bibitem[{Thrush et~al.(2022)Thrush, Jiang, Bartolo, Singh, Williams, Kiela, and Ross}]{thrush_and_ross2022winoground}
Tristan Thrush, Ryan Jiang, Max Bartolo, Amanpreet Singh, Adina Williams, Douwe Kiela, and Candace Ross. 2022.
\newblock Winoground: Probing vision and language models for visio-linguistic compositionality.
\newblock In \emph{CVPR}.

\bibitem[{Wang et~al.(2018)Wang, Singh, Michael, Hill, Levy, and Bowman}]{wang-etal-2018-glue}
Alex Wang, Amanpreet Singh, Julian Michael, Felix Hill, Omer Levy, and Samuel Bowman. 2018.
\newblock \href {https://doi.org/10.18653/v1/W18-5446} {{GLUE}: A multi-task benchmark and analysis platform for natural language understanding}.
\newblock In \emph{Proceedings of the 2018 {EMNLP} Workshop {B}lackbox{NLP}: Analyzing and Interpreting Neural Networks for {NLP}}, pages 353--355, Brussels, Belgium. Association for Computational Linguistics.

\bibitem[{Warstadt et~al.(2023{\natexlab{a}})Warstadt, Mueller, Choshen, Wilcox, Zhuang, Ciro, Mosquera, Paranjabe, Williams, Linzen, and Cotterell}]{warstadt-etal-2023-findings}
Alex Warstadt, Aaron Mueller, Leshem Choshen, Ethan Wilcox, Chengxu Zhuang, Juan Ciro, Rafael Mosquera, Bhargavi Paranjabe, Adina Williams, Tal Linzen, and Ryan Cotterell. 2023{\natexlab{a}}.
\newblock \href {https://doi.org/10.18653/v1/2023.conll-babylm.1} {Findings of the {B}aby{LM} challenge: Sample-efficient pretraining on developmentally plausible corpora}.
\newblock In \emph{Proceedings of the BabyLM Challenge at the 27th Conference on Computational Natural Language Learning}, pages 1--34, Singapore. Association for Computational Linguistics.

\bibitem[{Warstadt et~al.(2023{\natexlab{b}})Warstadt, Mueller, Choshen, Wilcox, Zhuang, Ciro, Mosquera, Paranjabe, Williams, Linzen, and Cotterell}]{conll-2023-babylm}
Alex Warstadt, Aaron Mueller, Leshem Choshen, Ethan Wilcox, Chengxu Zhuang, Juan Ciro, Rafael Mosquera, Bhargavi Paranjabe, Adina Williams, Tal Linzen, and Ryan Cotterell, editors. 2023{\natexlab{b}}.
\newblock \href {https://aclanthology.org/2023.conll-babylm.0} {\emph{Proceedings of the BabyLM Challenge at the 27th Conference on Computational Natural Language Learning}}. Association for Computational Linguistics, Singapore.

\bibitem[{Warstadt et~al.(2019)Warstadt, Parrish, Liu, Mohananey, Peng, Wang, and Bowman}]{Warstadt2019BLiMPTB}
Alex Warstadt, Alicia Parrish, Haokun Liu, Anhad Mohananey, Wei Peng, Sheng-Fu Wang, and Samuel~R. Bowman. 2019.
\newblock \href {https://api.semanticscholar.org/CorpusID:208527435} {Blimp: The benchmark of linguistic minimal pairs for english}.
\newblock \emph{Transactions of the Association for Computational Linguistics}, 8:377--392.

\bibitem[{Zhang et~al.(2015)Zhang, Goyal, Summers-Stay, Batra, and Parikh}]{vqav1}
Peng Zhang, Yash Goyal, Douglas Summers-Stay, Dhruv Batra, and Devi Parikh. 2015.
\newblock \href {https://api.semanticscholar.org/CorpusID:6733279} {Yin and yang: Balancing and answering binary visual questions}.
\newblock \emph{2016 IEEE Conference on Computer Vision and Pattern Recognition (CVPR)}, pages 5014--5022.

\end{thebibliography}

\appendix



\end{document}